# An Image-Based Path Planning Algorithm Using a UAV Equipped with Stereo Vision

Selim Ahmet IZ and Mustafa UNEL
Faculty of Engineering and Natural Sciences, Sabanci University, Istanbul, Turkey
{izselim, munel}@sabanciuniv.edu

*Abstract*—This paper presents a novel image-based path planning algorithm that was developed using computer vision techniques, as well as its comparative analysis with well-known deterministic and probabilistic algorithms, namely A* and Probabilistic Road Map algorithm (PRM). The terrain depth has a significant impact on the calculated path safety. The craters and hills on the surface cannot be distinguished in a two-dimensional image. The proposed method uses a disparity map of the terrain that is generated by using a UAV. Several computer vision techniques, including edge, line and corner detection methods, as well as the stereo depth reconstruction technique, are applied to the captured images and the found disparity map is used to define candidate way-points of the trajectory. The initial and desired points are detected automatically using ArUco marker pose estimation and circle detection techniques. After presenting the mathematical model and vision techniques, the developed algorithm is compared with well-known algorithms on different virtual scenes created in the V-REP simulation program and a physical setup created in a laboratory environment. Results are promising and demonstrate effectiveness of the proposed algorithm.

## I. INTRODUCTION

The path planning problem is based on optimization problems, which have been a popular research topic since the middle of the 1960s. Many deterministic and probabilistic techniques have been developed, but only a few of them have achieved a sufficient level of efficiency to be used in robotic systems [1]. Many of the prior studies focused on finding the shortest path between two points on a plane without going into details of how the map was generated. The image based path planning algorithms allowed the use of actual images of the terrain to determine where obstacles and the computed path would be located [2]. The implementation of this method with onboard cameras has significantly improved mobile robot navigation tasks.

Several image-based approaches exist in the literature for determining the optimal path by utilizing simple image characteristics [3] [4] or only depth information with grids [5] [6] [7]. In this context, there has been an increase in the usage of stereo vision depth reconstruction in path planning algorithms, particularly in the last century's quadrupeds [8], rovers [9], and off-road vehicles [10]. However, the depth information utilized to compute the local path in most of these studies. In some of the papers, the global path was generated by using the most efficient and favored path planning algorithms, such as A*, PRM, and RRT*, run on the Occupancy Grid of the map, which is the binary representation of the original terrain image [11] [12]. Hence, these algorithms cannot ensure that they will provide the correct path when the terrain's depth is unknown. Therefore, particularly in the recent decade, numerous image-based path planning algorithms have been created that use topographic maps (2.5D maps) to solve the unknown depth problem [13] [14], or the depth reconstruction approaches have begun to be used to determine the next step of the robot on the surface [8].

In this study, a novel image-based path planning algorithm based on computer vision techniques is developed and used to determine a global path between an unmanned ground vehicle (UGV) and a desired destination. The images of the terrain are taken by a UAV equipped with a stereo camera system, and the proposed algorithm is powered by topographic map data that permits the recognition of hills and pits that cannot be identified in a single image [15]. In the earlier studies, the stereo vision was typically utilized in local path planning operations to identify the distance between obstacles and the camera for a collision-free trajectory. According to the research that has been conducted, stereo depth reconstruction and computer vision techniques are not employed to compute global path in unknown environments. By generating a global path, the proposed algorithm provides a solution to the challenges discussed in the previous paragraphs. The obtained results are demonstrated in V-REP based virtual test scenes (Figure 1) and laboratory-created actual test scene (Figure 2). The contributions of the paper can be summarized as follows:

- The path between initial and desired points, which are detected automatically, is estimated using surface depth and terrain image features.
- Using stereo depth reconstruction to compute a global path that avoids negative and positive obstacles.

The paper is organized as follows: In Section II, the mathematical background of the proposed path planning algorithm and the implemented vision techniques are described. In Section III, the outputs of the path planning algorithms are compared and reviewed. Finally, the paper concludes with some remarks and indicates possible future directions.

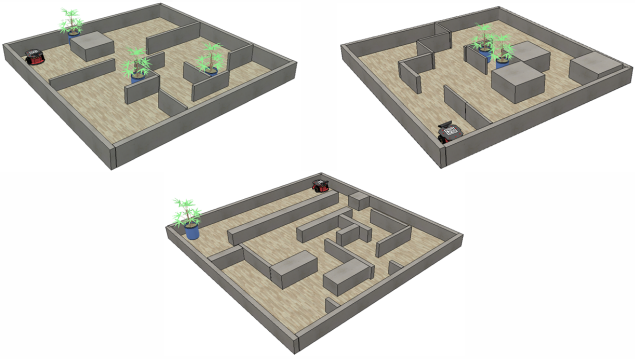

Fig. 1. Test Scenes in V-REP Simulation Program.

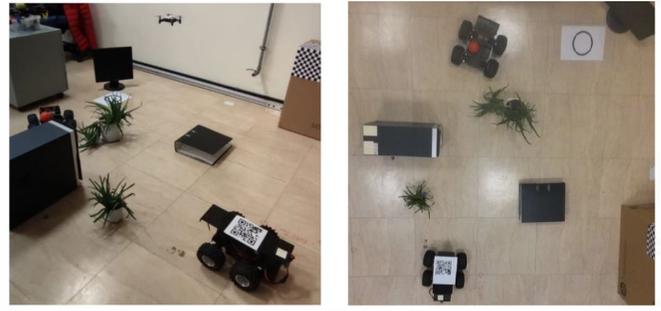

Fig. 2. Test Scene in Laboratory Environment.

## II. METHODOLOGY

In order to develop a global image-based path planning algorithm, the images that include both of the initial and desired points as well as the obstacles in the environment are required. In this study, a UAV outfitted with a stereo camera system is used to capture the images of the terrains and the stereo vision method is employed to compute depth information. Several computer vision techniques are implemented on acquired images to obtain the image characteristics necessary for drawing a path. Fast corner features are extracted from both 2D images i.e. one of the stereo pairs, and the disparity map obtained by stereo vision. After locating the detected corners in a single image, the way-points are chosen between them by comparing their pixel values with those of the initial and desired points. The unselected corners are eliminated and the number of way-point candidates is decreased. This elimination prevents the route from passing through a negative or positive obstacle, and reduces the computational load.

Although the detected corners are distinguished from the images, the path planning algorithms do not know which corner corresponds to which objects; hence, it is necessary to merge these corners with lines to differentiate the object edges from the path. This stage allows us to segment obstacles using Hough line detection algorithm. During the process, the algorithm thresholds that define the maximum and minimum length specifications are organized in accordance with the UGV's dimensions so that it can fit between various objects on the surface. To eliminate noisy outputs, the corner and line detection algorithms are applied to the original image after implementing edge detectors with smoothing capabilities.

In addition, the initial and desired points of the trajectory are automatically determined by the ArUco marker [16] and circle detection algorithms. The marker is used to detect the initial point and the orientation of the ground vehicle, and as shown in Figure 3, the initial point is positioned on top of the UGV with a blue dot. The desired point is designated by placing a circle in a similar manner, and the centers of the marker and the circle are given as the initial and desired points, respectively.

The stereo depth reconstruction technique is another crucial procedure for the developed path planning algorithm. The right and left images are used as input, and the hyperparameters $\omega$, which indicates the maximum limit of the color bar, and window size are adjusted through a series of tests to produce the optimal disparity maps. The obtained disparity maps indicate in different colors which item is closer to the camera or the image capture point, and the pixel information of every single features is located in the final column of the feature matrix as $P = \begin{bmatrix} x & y & I(p) \end{bmatrix}$. When the algorithm determines the depth, it assigns the same hue to the objects of equal height. Therefore, the way-point candidates of the path are picked by reapplying the corner detection algorithm with a pixel value condition in order to avoid positive and negative terrain obstacles. This method neither assigns points arbitrarily, as in probabilistic techniques, nor calculates all points in the image individually, making it easy to determine the shortest trajectory in a flat or low-slope location in a short amount of time.

The next step is to design the path planning algorithm utilizing detected corner points between the initial and desired locations. During the process, it is intended that the trajectory lines will not intersect the lines representing the object's edges. The necessary information for each of these techniques is provided below.

### A. The Proposed Path Planning Algorithm

In the developed algorithm the following computer vision (feature extraction) techniques were utilized, and the path was calculated, accordingly.

*a) Edge Detection:* Edges appear at the boundary between two distinct image regions. These algorithms use a number of mathematical techniques to find edges, which are curves in a digital image where the image brightness abruptly changes or, more formally, contains discontinuities. One of the first order derivative edge detection filters Canny (1st order) is employed as a pre-processing step in this context due to the excessively noisy results produced by the corner and line detection methods. This filter was specifically chosen due to its smoothing property.

*b) Corner Detection:* The corner detection algorithms follow similar steps until the completion of the $H$ matrix computation, which is a matrix whose elements are computed by different combinations of image gradients. The eigenvalues of $H$ matrix and the $f$ value, which can be computed as $f = det(H)/trace(H)$, are used in comparison with predefined thresholds. The FAST corner detection technique is one of the quickest feature extraction techniques, such as the difference of Gaussians (DoG) employed by the SIFT, SUSAN, and Harris detectors. In its operating principle, a 16-pixel circle, also known as the Bresenham circle, is utilized to classify the candidate point $p$ that remains in the circle's center [17]. Each pixel in the circle is labeled with an integer from 1 to 16 in a clockwise direction, and the corner point is identified by comparing the pixels of the two points. In this study, the FAST feature technique is favored because it produces corners with less noise than the other techniques.

*c) Line Detection:* The line detection algorithm is one of the most crucial steps of the proposed method, as it clarifies the obstacle's borders. Despite the fact that the corners of the obstacles were identified in earlier steps, they provide no information regarding the boundary locations and directions. It is important to merge the corner points according to a rule in order to give them meaning. Due to the presence of several noisy corners in the input images, it is challenging to identify the right matches that define the obstacle edges. In this stage, the Hough line detector is therefore employed by specifying particular conditions for the head and tail nodes of the lines. In the implementation, nodes are assigned a distance condition, and the whole nodes inside the circular area whose radius smaller than the defined reference distance are merged.

*d) Marker Detection - ArUco Marker:* In the developed algorithm, a marker is placed on top of the UGV to automatically detect the initial point without manual intervention [16]. Since this technique provides both the position and orientation of the object, ArUco marker is chosen. Figure 3, depicts the initial point locations for all test scenes, which were determined using the marker mounted to the UGV. In these figures, the centroid of the detected marker is shown by a red dot, and the beginning point, which is denoted by a small blue square, is automatically found based on the red dot.

*e) Circle Detection:* The Hough circle detection technique is used to locate the target point, which is likewise denoted by a circle in the scene similar to the usage goal of the marker detection method. The Circle Hough Transform (CHT) is a fundamental feature extraction technique used in digital image processing to detect incomplete circles in images. The candidates for circles are generated by "voting" in the Hough parameter space and then selecting local maxima from an accumulator matrix. A circle can be characterized in a two-dimensional space as $(x - a)^2 + (y - b)^2 = r^2$, where (a, b) is the circle's center and r is its radius. In 3D space, the circle parameters can be identified via the intersection of several conic surfaces created by 2D circle points. This procedure consists of two steps. First, the radius is fixed, and then the optimal center of circles in a 2D parameter space is determined. In the second step, the ideal radius in one-dimensional parameter space is computed. Since the algorithm can recognize another circular shape and assign it as a target point, the maximum/minimum radius of circle and sensitivity value are defined as hyperparameters: $Rmax = 40$, $Rmin = 10$, $Sensitivity = 0.9$. Figure 3 illustrates the acquired outcomes.

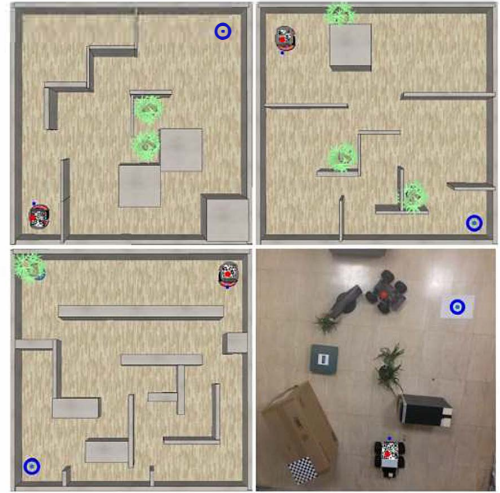

Fig. 3. ArUco marker (Initial Point Detection) and Circle Detection (Desired Point Detection) Left Top Image: Scene-1(V-rep), Right Top Image: Scene-2(V-rep), Left Down Image: Scene-3(V-rep), Right Down Image: Scene-4 (Real Setup)

*f) Stereo Depth Reconstruction:* The terrain's depth information is one of the pillars of the created path planning algorithm. Stereo vision algorithm makes it possible to assess the trajectory's slope and choose a direction with a gentle gradient. The initial stage in determining depth from stereo vision is breaking the images into positive and negative patches by dividing the right and left images into patches. After the patches are separated according to their shapes, sizes, and relative geometries, the matching operation initiates. There are various ways for matching stereo images, however they can be grouped primarily into two main categories: area-based correlation techniques and feature-based techniques. In the correlation method, the right and left images are scanned pixel by pixel to identify relationships between the small rectangular portions. In contrast, feature-based approaches strive to detect a certain class of feature between two images. After matching pixels from two input images, all detected features are stored in a map known as a disparity map. If the camera lens parameters are known, such as focal length $f$ and baseline distance $T = 2l$, the depth can be calculated from the disparity map. As can be seen in the following equations, the position and specifications of the stereo camera system, such as the properties of camera lens and the distance of the camera pairs, are essential for calculating depth. Additionally, it is vital to capture adequate light in an appropriate environment while avoiding reflection, transparency, and mirror surfaces for high quality results.

$$d = x_L - x_R = f(\frac{x_p + l}{z_p}) - (\frac{x_p - l}{z_p}) = \frac{2fl}{z_p}$$

$$\Rightarrow z_p = \frac{2fl}{d}$$

The hyperparameters of the stereo depth estimation algorithm have a significant impact on the output. These hyperparameters, referred as $\omega$, that defines the color map pixel value interval, and the window size utilized for pixel scanning. High values of the window size effects the disparity map quality negatively, but the processing time decreases. During the experiments, the shadow effect on the results was minimized by comparing the two input images with different windows widths. Figure 4 shows the disparity maps of the test scenes and the associated hyperparameter values.

As seen in Figure 4, when the same hyperparameters are used, the disparity map of the actual scene does not seem as good as those of the other scenes. There are two primary causes for this problem. The first is the image pairs that have been used: Since the autonomous UAV that is used to capture images of the actual environment has a monocular onboard camera, the right and left images are taken after moving the UAV along an axis. The transition of the UAV is taken as baseline and the disparity map is generated under this condition. However, since the image pairs in stereo vision technique should be aligned in the same direction, the unanticipated axis shifts that occur during the transition of the UAV negatively influences the quality of the captured image pairs and disparity map. The second reason is the impact on the environment. Due to the fact that the surface of the terrain reflects the indoor lighting, a distinct group of artificial colors appeared on the image. Despite the fact that the depth estimate algorithm did not produce a satisfactory result in a real-world scenario, the generated path planning algorithm operates without any issue since it takes into account the corner spots discovered from the disparity maps.

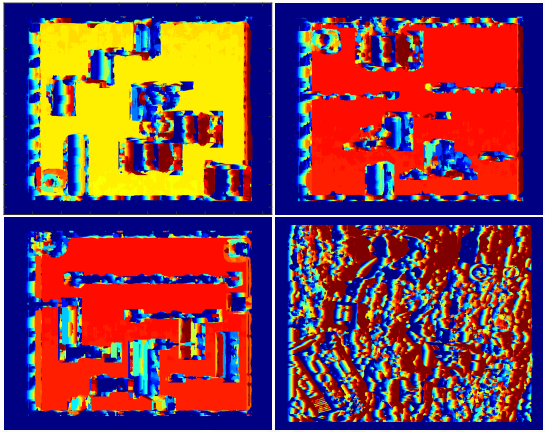

Fig. 4. Disparity Maps of All Test Scenes ($\omega$ =50,$WindowSize$=10) Left Top Image: Scene-1(V-rep), Right Top Image: Scene-2(V-rep), Left Down Image: Scene-3(V-rep), Right Down Image: Scene-4 (Real Setup).

*g) Line Intersection Algorithm:* The line intersection method is the most essential aspect of the developed path planning algorithm. After locating the corner points come from terrain image and the disparity map, the best corner points should be identified as way-points for use in building a trajectory by connecting them with straight lines. It is also necessary to prevent any intersections with other lines that are drawn by the Hough line detection method to detect the obstacle edges in the terrain. In accordance with this approach, a line-drawing method is developed along with a condition to avoid line intersection scenarios by utilizing simple geometrical and analytical data.

In Figure 5, a line intersection scenario is depicted. In this diagram, the black line indicates the shortest path between the initial and final positions when the line intersection requirement is deactivated. In addition, point $P_1$ represents a point on the trajectory, point $P_2$ indicates a location on one of the edge lines that intersects with the trajectory, and point $P_3$ represents the intersection point. Under these conditions, when the angles between the horizontal axis and the blue and black lines are referred to as $\theta_1$ and $\theta_2$, respectively, the line equations become:

$$L_1 = P_1 + \lambda_1 \left[ \begin{array}{c} cos(\theta_1) \\ sin(\theta_1) \end{array} \right], \quad L_2 = P_2 + \lambda_2 \left[ \begin{array}{c} cos(\theta_2) \\ sin(\theta_2) \end{array} \right]$$

The intersection point $P_3$ can be incorporated into the equations of $L_1$ and $L_2$:

$$P_3 = P_1 + \lambda_{1_{P_3}} \left[ \begin{array}{c} cos(\theta_1) \\ sin(\theta_1) \end{array} \right], \quad P_3 = P_2 + \lambda_{2_{P_3}} \left[ \begin{array}{c} cos(\theta_2) \\ sin(\theta_2) \end{array} \right]$$

These equations can be solved simultaneously as:

$$P_1 + \lambda_{1_{P_3}} \left[ \begin{array}{c} cos(\theta_1) \\ sin(\theta_1) \end{array} \right] = P_2 + \lambda_{2_{P_3}} \left[ \begin{array}{c} cos(\theta_2) \\ sin(\theta_2) \end{array} \right]$$

Assuming the matrix is invertible, the equation can be expressed as follows:

$$\left[ \begin{array}{c} \lambda_{1_{P_3}} \\ \lambda_{2_{P_3}} \end{array} \right] = \left[ \begin{array}{cc} cos(\theta_1) & -cos(\theta_2) \\ sin(\theta_1) & -sin(\theta_2) \end{array} \right]^{-1} (P_1 - P_2)$$

The coordinates of the intersection point can be determined by substituting one of the $\lambda$ values found in preceding step into one of the line equations specified in the first step. In this stage, the $\lambda$ values act an important role in the selection of the candidate way-points since their values between 0 and 1, refer to the intersection of the lines.

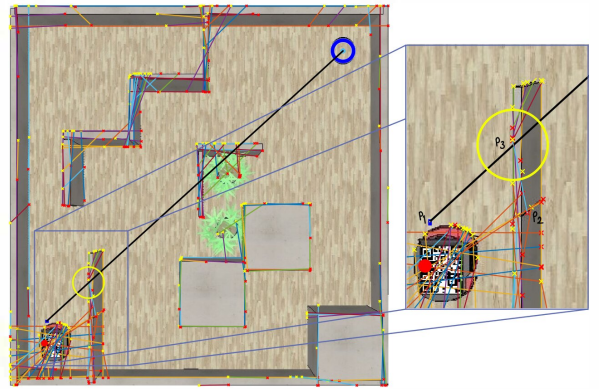

Fig. 5. Line Intersection Case in Scene 1.

## III. RESULTS AND DISCUSSION

The proposed path planning algorithm and the well-known probabilistic and deterministic algorithms which are PRM and A*, are implemented to the V-rep based simulation environment and the actual environment established in laboratory. During the experiment the hyperparameters of stereo depth reconstruction method are never changed and selected as $\omega = 50, WindowSize = 10$. Similarly, the sensitivity value of the circle detection algorithm has not been changed as well as the thresholds of the radius; and the same marker used top of the UGV for pose estimation. Throughout all of these experiments, a Lenovo E490 ThinkPad with 16GB of RAM is used as a work-station.

The initial and end points of the drawn lines by Hough line detection method, are indicated with red cross signs. Similarly, the corners detected on disparity map are represented with red asterisks, and the corners found from the original image are showed with green plus signs on the figures below, after made the elimination according to pixel values as explained in the Section II.

The acquired results indicate that the A* approach requires either longer processing time or stronger computational capacity, but the PRM algorithm does not always provide the optimal path since it employs random points, despite the fact that it locates the paths in a shorter amount of time. The computing time and energy demand of the PRM algorithm are proportional to the number of thrown way-points. In addition, neither of these two well-known path planning algorithm take the depth information of the ground into account. They exclusively utilize information derived from the occupancy grid form of terrain image. Therefore, it is not possible to identify hills or pits that are not visible in a single top view of image.

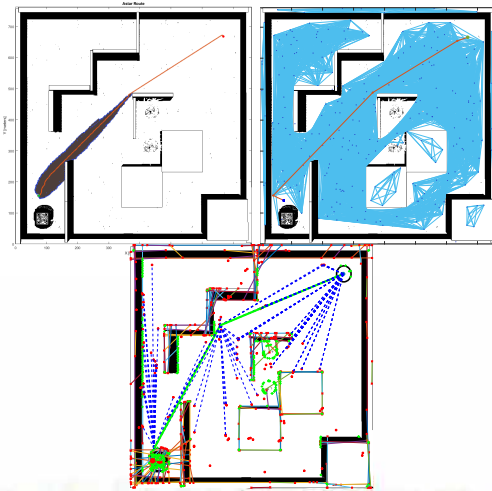

Fig. 6. Found Paths in the Scene-1 (V-rep: 759x763) Left Top Image:A*, Right Top Image: PRM, Down-Side Image: Proposed Algorithm.

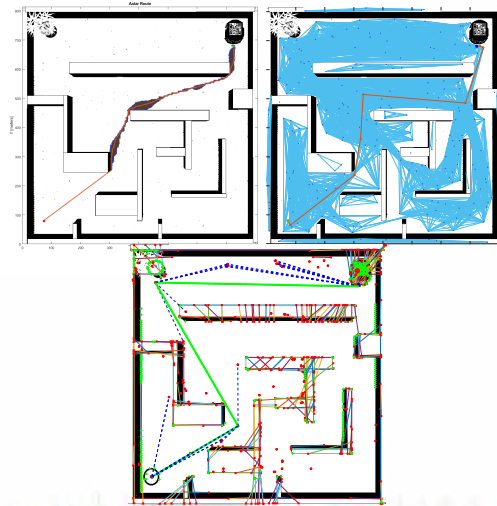

Fig. 8. Found Paths in the Scene-3 (V-rep: 808x814) Left Top Image:A*, Right Top Image: PRM, Down-Side Image: Proposed Algorithm.

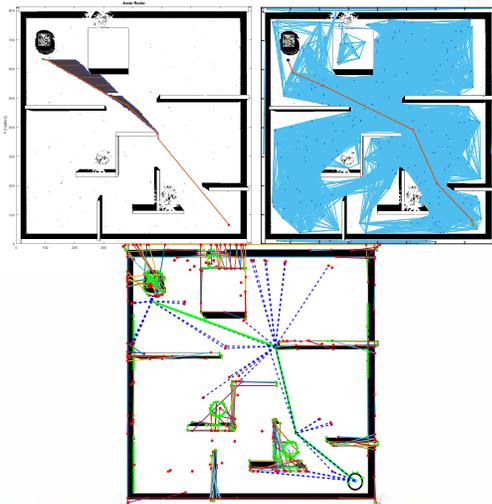

Fig. 7. Found Paths in the Scene-2 (V-rep: 808x814) Left Top Image:A*, Right Top Image: PRM, Down-Side Image: Proposed Algorithm.

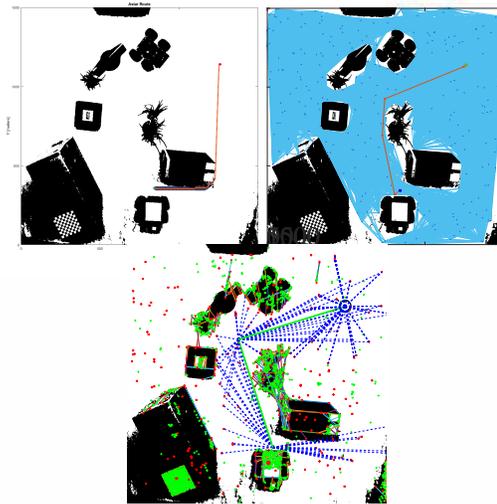

Fig. 9. Found Paths in the Scene-4 (Real scene: 1500x1500) Left Top Image:A*, Right Top Image: PRM, Down-Side Image: Proposed Algorithm.

Consequently, the generated path of proposed algorithm is comparable to the outputs of the PRM algorithm in all cases, but more optimal due to the way-point selection. Although the PRM and proposed algorithm computation time differ by only a few seconds, the A* algorithm has fallen significantly behind these two methods. Moreover, the developed algorithm is different than the others in that it concurrently considers the image characteristics and the terrain's depth information.

As a result, the developed algorithm only computed a distinct path in test scene 3, as it takes into account the width and length of the UGV to select healthier way-points where the UGV can fit. According to test scenes listed in the tables below, the computation time of the path planning algorithms is provided.

TABLE I
PATH PLANNING ALGORITHM COMPUTATION TIME RESULTS IN SCENE-1

| Algorithm Name | Computation Time (seconds) |
|---|---|
| PRM | 2.520640 |
| A* | 123.036743 |
| Proposed Algorithm | 1.720869 |

TABLE II
PATH PLANNING ALGORITHM COMPUTATION TIME RESULTS IN SCENE-2

| Algorithm Name | Computation Time (seconds) |
|---|---|
| PRM | 2.786833 |
| A* | 124.769130 |
| Proposed Algorithm | 1.613129 |

TABLE III
PATH PLANNING ALGORITHM COMPUTATION TIME RESULTS IN SCENE-3

| Algorithm Name | Computation Time (seconds) |
|---|---|
| PRM | 2.695357 |
| A* | 157.781910 |
| Proposed Algorithm | 1.873277 |

TABLE IV
PATH PLANNING ALGORITHM COMPUTATION TIME RESULTS IN SCENE-4

| Algorithm Name | Computation Time (seconds) |
|---|---|
| PRM | 4.889430 |
| A* | 109.420019 |
| Proposed Algorithm | 3.379621 |

## IV. CONCLUSION

In this paper, a novel path planning algorithm is developed by using computer vision techniques, and its performance is compared to that of well-known probabilistic and deterministic path planning algorithms. The proposed algorithm utilizes the terrain depth map obtained through stereo-vision depth reconstruction method, and draws optimal path using several computer vision techniques. After comparing the computed paths on created V-REP and physical test scenes, we have demonstrated that the proposed path planning algorithm is faster and safer than A* and PRM algorithms in several scenarios. As a future work, we plan to utilize camera calibration to get metric information in 3D environment [18] and avoid dynamic obstacles [19] as well as conduct outdoor GPS based tests.

REFERENCES

[1] Ferguson, D., Likhachev, M. and Stentz, A., 2005, June. A guide to heuristic-based path planning. In Proceedings of the international workshop on planning under uncertainty for autonomous systems, international conference on automated planning and scheduling (ICAPS) (pp. 9-18).
[2] Noborio, H. and Nishino, Y., 2001, May. Image-based path-planning algorithm on the joint space. In Proceedings 2001 ICRA. IEEE International Conference on Robotics and Automation (Cat. No. 01CH37164) (Vol. 2, pp. 1180-1187). IEEE.
[3] Yonetani, R., Taniai, T., Barekatain, M., Nishimura, M. and Kanezaki, A., 2021, July. Path planning using neural a* search. In International Conference on Machine Learning (pp. 12029-12039). PMLR.
[4] Thoma, J., Paudel, D.P., Chhatkuli, A., Probst, T. and Gool, L.V., 2019. Mapping, localization and path planning for image-based navigation using visual features and map. In Proceedings of the IEEE/CVF Conference on Computer Vision and Pattern Recognition (pp. 7383-7391).
[5] Folsom, L., Ono, M., Otsu, K. and Park, H., 2021. Scalable information-theoretic path planning for a rover-helicopter team in uncertain environments. International Journal of Advanced Robotic Systems, 18(2), p.1729881421999587.
[6] Peng, M., Di, K., Wang, Y., Wan, W., Liu, Z., Wang, J. and Li, L., 2021. A Photogrammetric-Photometric Stereo Method for High-Resolution Lunar Topographic Mapping Using Yutu-2 Rover Images. Remote Sensing, 13(15), p.2975.
[7] Viseras, A., Shutin, D. and Merino, L., 2019. Robotic active information gathering for spatial field reconstruction with rapidly-exploring random trees and online learning of Gaussian processes. Sensors, 19(5), p.1016.
[8] Bazeille, S., Barasuol, V., Focchi, M., Havoutis, I., Frigerio, M., Buchli, J., Caldwell, D.G. and Semini, C., 2014. Quadruped robot trotting over irregular terrain assisted by stereo-vision. Intelligent Service Robotics, 7(2), pp.67-77.
[9] Giubilato, R., Vayugundla, M., Schuster, M.J., Stürzl, W., Wedler, A., Triebel, R. and Debei, S., 2020. Relocalization with submaps: Multi-session mapping for planetary rovers equipped with stereo cameras. IEEE Robotics and Automation Letters, 5(2), pp.580-587.
[10] Joseph, L. and Mondal, A.K. eds., 2021. Autonomous Driving and Advanced Driver-assistance Systems (ADAS): Applications, Development, Legal Issues, and Testing. CRC Press.
[11] Niu, H., Lu, Y., Savvaris, A. and Tsourdos, A., 2018. An energy-efficient path planning algorithm for unmanned surface vehicles. Ocean Engineering, 161, pp.308-321.
[12] Khanmirza, E., Haghbeigi, M., Nazarahari, M. and Doostie, S., 2017, October. A comparative study of deterministic and probabilistic mobile robot path planning algorithms. In 2017 5th RSI International Conference on Robotics and Mechatronics (ICRoM) (pp. 534-539). IEEE.
[13] Liu, Z., Di, K., Li, J., Xie, J., Cui, X., Xi, L., Wan, W., Peng, M., Liu, B., Wang, Y. and Gou, S., 2020. Landing site topographic mapping and rover localization for Chang'e-4 mission. Science China Information Sciences, 63(4), pp.1-12.
[14] Li, J., Li, X. and Yu, L., 2018, May. Multi-UAV cooperative coverage path planning in plateau and mountain environment. In 2018 33rd youth academic annual conference of Chinese association of automation (YAC) (pp. 820-824). IEEE.
[15] Allan, M., Mcleod, J., Wang, C., Rosenthal, J.C., Hu, Z., Gard, N., Eisert, P., Fu, K.X., Zeffiro, T., Xia, W. and Zhu, Z., 2021. Stereo correspondence and reconstruction of endoscopic data challenge. arXiv preprint arXiv:2101.01133.
[16] Marut, A., Wojtowicz, K. and Falkowski, K., 2019, June. ArUco markers pose estimation in UAV landing aid system. In 2019 IEEE 5th International Workshop on Metrology for AeroSpace (MetroAeroSpace) (pp. 261-266). IEEE.
[17] Wang, Y., Li, Y., Wang, J., Lv, H. and Yang, Z., 2022. A Target Corner Detection Algorithm Based on the Fusion of FAST and Harris. Mathematical Problems in Engineering, 2022.
[18] Rodríguez-Quiñonez, J.C., Sergiyenko, O., Flores-Fuentes, W., Rivas-Lopez, M., Hernandez-Balbuena, D., Rascón, R. and Mercorelli, P., 2017. Improve a 3D distance measurement accuracy in stereo vision systems using optimization methods' approach. Opto-Electronics Review, 25(1), pp.24-32.
[19] Lin, H.Y. and Peng, X.Z., 2021. Autonomous quadrotor navigation with vision based obstacle avoidance and path planning. IEEE Access, 9, pp.102450-102459.